\documentclass[runningheads]{llncs}
\usepackage[T1]{fontenc}
\usepackage{graphicx}
\usepackage{cite}
\usepackage{amsmath,amssymb,amsfonts}
\usepackage{algorithmic}
\usepackage{algorithm}
\usepackage{graphicx}
\usepackage{textcomp}
\usepackage{xcolor}
\usepackage{float}
\restylefloat{table}
\usepackage{lscape}
 \usepackage[numbers]{natbib}
\usepackage{graphicx}
\usepackage{caption}
\usepackage{subcaption}
\usepackage{booktabs,makecell}
\usepackage{multirow}
\usepackage{pdflscape}
\usepackage{caption}
\usepackage{hyperref}
\usepackage{rotating}
\usepackage[export]{adjustbox}
\usepackage{comment}
\usepackage{amsmath}
\usepackage{xcolor}
\usepackage{tikz}
\usetikzlibrary{shapes,arrows}
\usepackage{placeins}
\usepackage{graphics}
\usepackage{xcolor}
\usepackage{mathrsfs}

\def\BibTeX{{\rm B\kern-.05em{\sc i\kern-.025em b}\kern-.08em
    T\kern-.1667em\lower.7ex\hbox{E}\kern-.125emX}}

\begin{document}

\title{GL-TSVM: A robust and smooth twin support vector machine with guardian loss function}

\author{Mushir Akhtar\inst{1} \and
M. Tanveer\inst{1}\thanks{ \noindent Corresponding Author} \and
Mohd. Arshad\inst{1}
}
\authorrunning{Akhtar et al.}
\institute{Indian Institute of Technology Indore, Simrol, Indore, India 
\email{\{phd2101241004,mtanveer,arshad\}@iiti.ac.in}}
\maketitle              % typeset the header of the contribution
\begin{abstract}
Twin support vector machine (TSVM), a variant of support vector machine (SVM), has garnered significant attention due to its $3/4$ times lower computational complexity compared to SVM. However, due to the utilization of the hinge loss function, TSVM is sensitive to outliers or noise. To remedy it, we introduce the guardian loss (G-loss), a novel loss function distinguished by its asymmetric, bounded, and smooth characteristics. We then fuse the proposed G-loss function into the TSVM and yield a robust and smooth classifier termed GL-TSVM. Further, to adhere to the structural risk minimization (SRM) principle and reduce overfitting, we incorporate a regularization term into the objective function of GL-TSVM. To address the optimization challenges of GL-TSVM, we devise an efficient iterative algorithm. The experimental analysis on UCI and KEEL datasets substantiates the effectiveness of the proposed GL-TSVM in comparison to the baseline models. Moreover, to showcase the efficacy of the proposed GL-TSVM in the biomedical domain, we evaluated it on the breast cancer (BreaKHis) and schizophrenia datasets. The outcomes strongly demonstrate the competitiveness of the proposed GL-TSVM against the baseline models. The supplementary file, along with the source code for the proposed GL-TSVM model, is publicly accessible at \url{https://github.com/mtanveer1/GL-TSVM}.

\end{abstract}

\section{Introduction}
Support vector machine (SVM) \cite{cortes1995support}, a kernel-based method, has been extensively researched over the last two decades, particularly in the realm of pattern recognition. It is rooted in the concept of structural risk minimization (SRM) and is derived from statistical learning theory (SLT), consequently having a solid theoretical foundation and demonstrating better generalization capabilities. Its wide-ranging applications span across diverse domains, including cancer diagnosis \cite{kumari2024diagnosis}, Alzheimer detection \cite{10619989}, and so forth.
% , EEG signal classification \cite{ganaie2023eeg}, and so forth.
\par
The key idea of SVM revolves around identifying two parallel hyperplanes with the maximum possible margin between them. This primarily involves solving a quadratic programming problem (QPP) whose complexity is proportional to the cube of the training dataset size. Twin SVM (TSVM) \cite{khemchandani2007twin}, a variant of SVM, tackles this problem by solving two smaller QPPs instead of one large QPP, thereby reducing computational costs by approximately 75\% compared to traditional SVM methods. This significant computational efficiency has garnered considerable attention from the research community, leading to extensive studies aimed at enhancing the performance of TSVM. For instance, \citet{kumar2009least} proposed the least squares TSVM, which stands out for its simplicity and efficiency in binary classification tasks. Additionally, \citet{shao2011improvements} introduced another improved variant, the twin bounded SVM, which boosts the generalization capabilities of TSVM through the incorporation of structural risk minimization term. Besides, numerous researchers have developed several TSVM variants to further enhance its performance across different applications. As an example, to tackle the imbalance problem, \citet{ganaie2022large} proposed the large-scale fuzzy least squares TSVM. To develop a robust and sparse variant of TSVM, \citet{tanveer2015robust} reformulated the classical TSVM by incorporating a regularization technique and proposed an exact 1-norm linear programming formulation for TSVM. To address the noise sensitivity of TSVM, \citet{tanveer2022large} proposed the large-scale pin-TSVM by utilizing the pinball loss function. Both of the aforementioned algorithms eliminate the requirement of matrix inversion, making them suitable for large-scale problems. To delve deeper into the development of TSVM models, readers can refer to \cite{tanveer2019comprehensive, tanveer2022comprehensive}.
\par
Despite several strengths of TSVM, it still has opportunities for improvement. One key area is its sensitivity to outliers or noise, which stems from the unbounded escalation of the hinge loss function, leading to excessively high losses for samples located far from the proximal hyperplane \cite{wang2020comprehensive}. Also, the hinge loss function solely imposes penalties on misclassified samples and neglects the contribution of the correctly classified samples. However, the influence of samples from distinct classes, positioned on either side of the hyperplane, on the decision hyperplane differs based on their respective locations. Various researchers have conducted extensive investigations to reduce the susceptibility of TSVM to noise and outliers \cite{ganaie2022knn, si2023symmetric}. Among these approaches, designing a robust loss function has emerged as a crucial focus. Recent advancements have introduced several robust loss functions for TSVM, such as the symmetric LINEX loss function \cite{si2023symmetric}, pinball loss function \cite{tanveer2022large}, correntropy-induced loss function \cite{zheng2021ctsvm}, Huber loss function \cite{borah2020functional}, and others. Despite their contributions to enhancing the robustness of TSVM, these methods still present notable limitations. The symmetric LINEX loss and Huber loss functions, while innovative, suffer from symmetry and unbounded growth. Their unbounded nature makes them vulnerable to extreme values, thereby increasing sensitivity to outliers or noise. Additionally, their symmetric design treats samples on either side of the proximal hyperplane equally, potentially ignoring the differing influences these samples exert on the decision boundary. The correntropy-induced loss, though bounded, maintains symmetry, similarly equalizing the importance of samples regardless of their positioning. Conversely, the pinball loss function introduces asymmetry but lacks boundedness, failing to cap the impact of extreme data points. In essence, although these loss functions mark significant strides in the robustness of TSVM, their inherent limitations highlight the ongoing need for further research. Developing more effective and versatile loss functions is essential to advancing robust machine learning models, capable of better handling outliers and noise, thereby ensuring more reliable and accurate predictions.
\par
Taking motivation from prior research, in this paper, we develop a robust loss function, named guardian loss (G-loss). It is meticulously designed to possess asymmetric, bounded, and smooth characteristics. Then, we amalgamate the proposed  G-loss function into TSVM and introduce a robust and smooth classifier termed GL-TSVM. The main contributions of this paper can be outlined as follows:
\begin{enumerate}
    \item To shield TSVM against outliers or noise, we introduce the guardian loss function (G-loss), a novel approach characterized by its asymmetry, boundedness, and smoothness. The asymmetric feature of G-loss enables the assignment of distinct penalties to distinct samples based on their location with respect to the proximal hyperplane. The bounded nature allows for a strict limit on the maximum loss for data points with significant deviations, thereby mitigating the influence of noise or outliers. Furthermore, the smoothness property empowers the utilization of gradient-based algorithms for model optimization.
    \item We incorporate the G-loss function into TSVM and propose a novel robust and smooth classifier named GL-TSVM. Additionally, we devised an iterative algorithm to address the optimization problems of GL-TSVM.
    \item To employ the principle of structural risk minimization (SRM) and avoid the overfitting problem, we introduce a regularization term into the objective function of GL-TSVM. 
    \item We perform numerical evaluation on benchmark UCI and KEEL datasets from various domains. The outcomes reveal the superior performance of the proposed GL-TSVM against the baseline models.
    \item To evaluate the superiority of the proposed GL-TSVM in the biomedical realm, we conducted experiments using the breast cancer (BreaKHis) and schizophrenia datasets. The results provide compelling evidence of the proposed GL-TSVM applicability in the biomedical domain.
\end{enumerate}
The rest of this paper is structured as follows: The related works are discussed briefly in Section 2. Section 3 presents the proposed  G-loss function and provides the formulation of GL-TSVM.
% along with an analysis of the computational complexity.
Section 4 showcases the results of the experiments. Lastly, Section 5 concludes the paper with future directions.

\section{Related Work}
This section begins by defining the notations employed in this paper. Following this, we provide a brief overview of
some relevant loss functions. The formulation of TSVM is briefly discussed in Section S.I of the supplement file. 
% relevant literature, including TSVM and 
\subsection{Notations}
Consider the training set denoted as $\left\{x_i,y_i\right\}_{i=1}^l$, where $x_i \in \mathbb{R}^n$ represents the sample vector and $y_i \in\{-1,1\}$ signifies the corresponding class label. Let ${X}_{+}=\left({x}_1, \ldots, {x}_{l_{+}}\right)^{\top} \in \mathbb{R}^{l_{+} \times n}$ and ${X}_{-}=\left({x}_1, \ldots, {x}_{l_{-}}\right)^{\top} \in$ $\mathbb{R}^{l_{-} \times n}$ represent matrices containing positive and negative instances, where $l_{+}$ and $l_{-}$ denote the count of positive and negative instances, respectively, and $l=l_{+}+l_{-}$.
Further, $e_1$ and $e_2$ are identity vectors of appropriate size, and $I$ is the identity matrix of appropriate size.
\begin{figure*}
\centering
    \subcaptionbox{     \label{fig:first }} { %
      \includegraphics[width=0.48\textwidth,keepaspectratio]{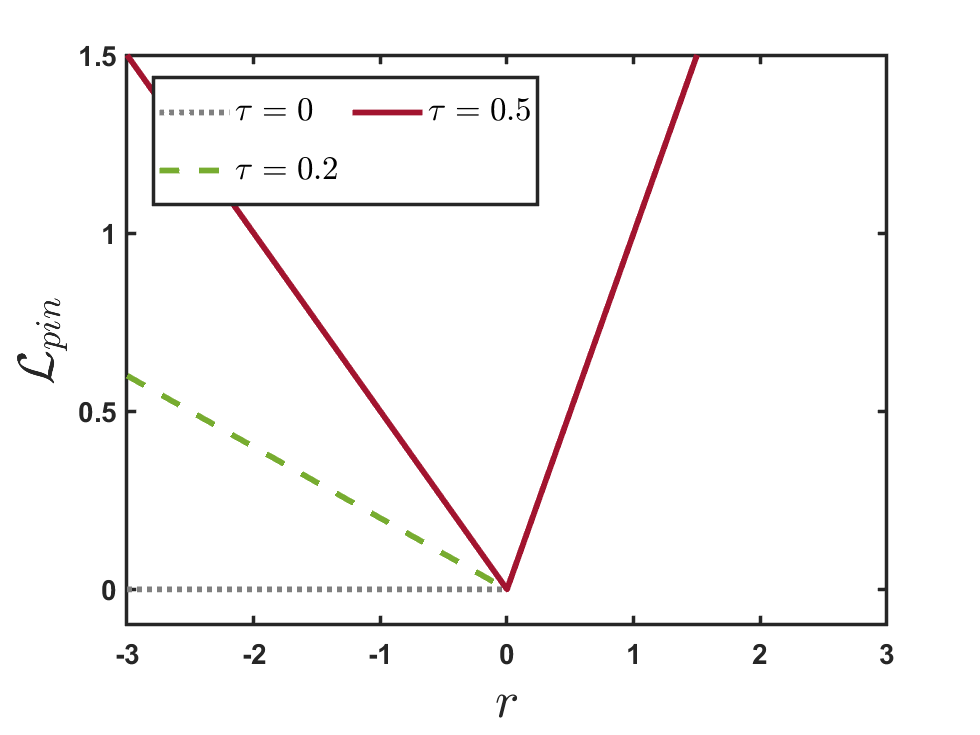}}
      \hfill
      \subcaptionbox{   \label{fig:second }} { %
      \includegraphics[width=0.48\textwidth,keepaspectratio]{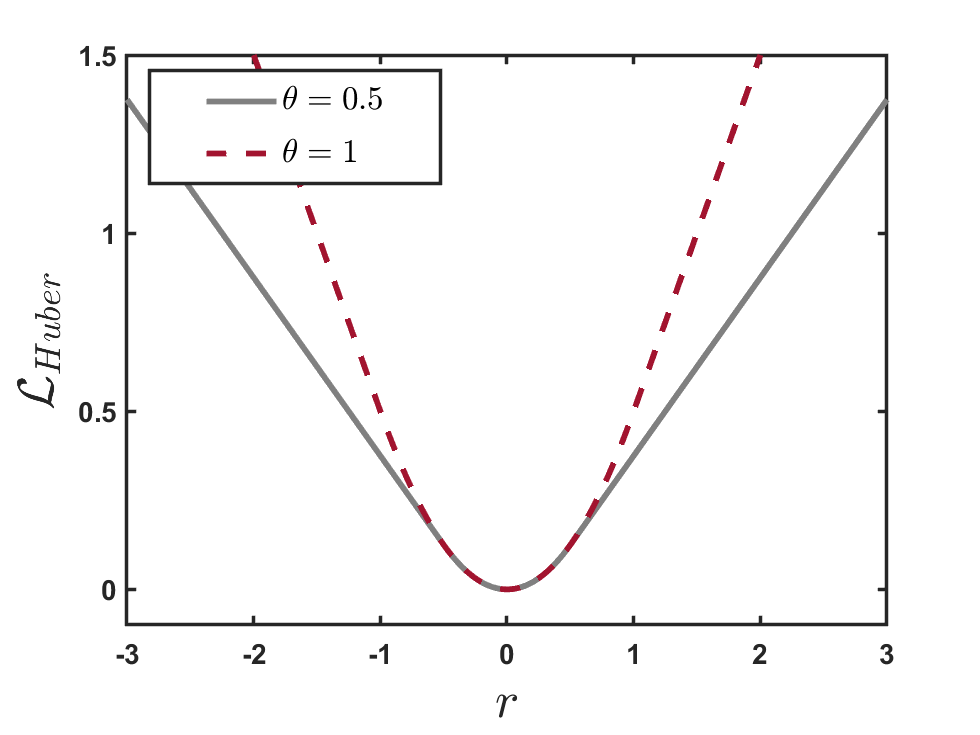}}
\\
      \subcaptionbox{  \label{fig:third}} { %
      \includegraphics[width=0.48\textwidth,keepaspectratio]{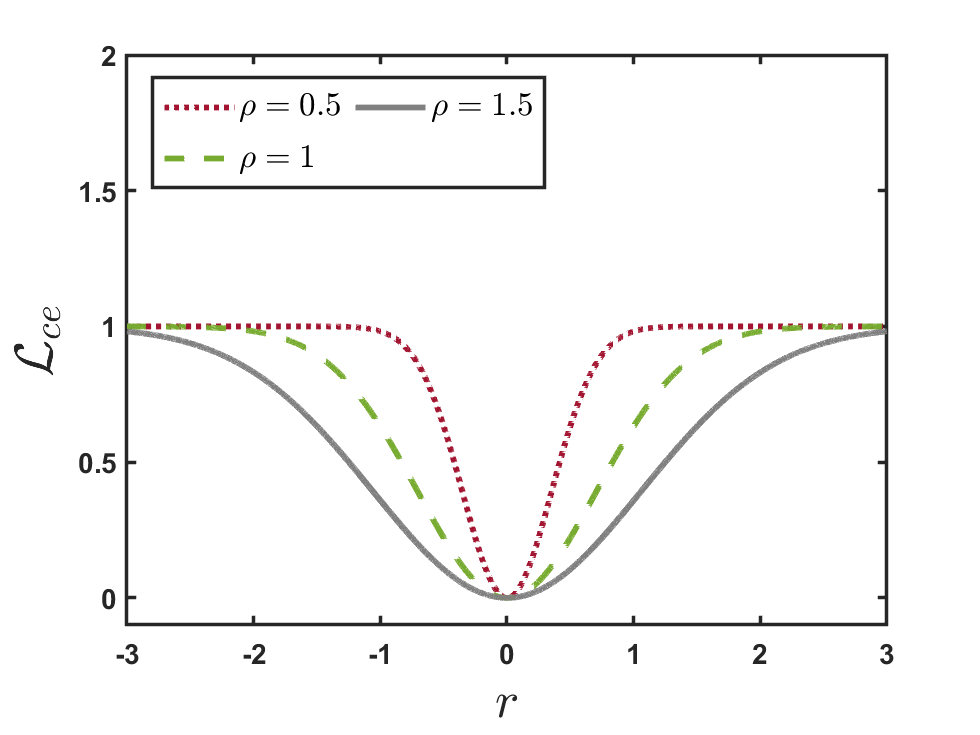}}
\hfill     
      \subcaptionbox{  \label{fig:four}} { %
      \includegraphics[width=0.48\textwidth,keepaspectratio]{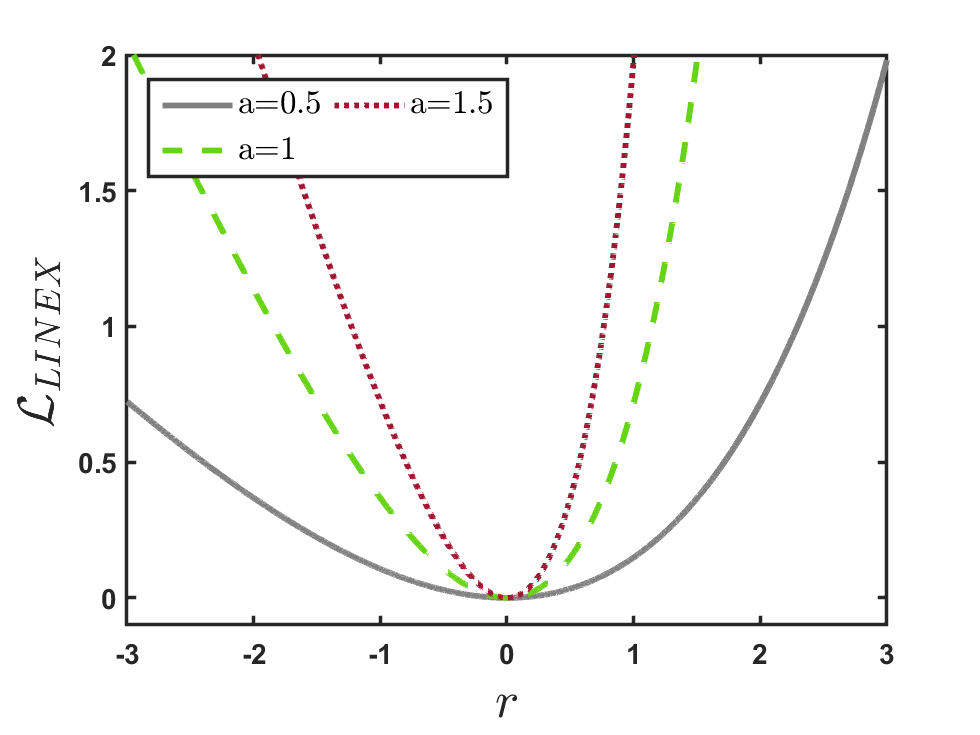}}
      \\
      \subcaptionbox{   \label{fig:fifth }} { %
      \includegraphics[width=0.55\textwidth,keepaspectratio]{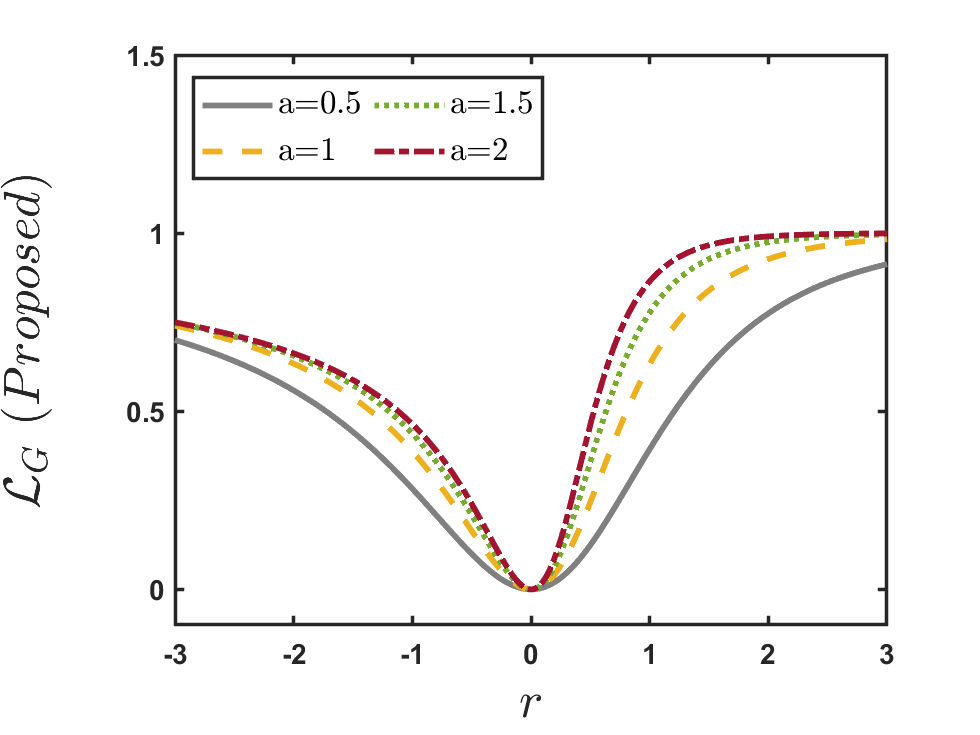}}
      \caption{Visual illustration of baseline and proposed G-loss function. (a) Pinball loss function with $\tau=0$, $\tau=0.2$, and $\tau=0.5$. (b) Huber loss function with $\theta=0.5$ and $\theta=1$. (c) Correntropy-induced loss function with $\rho=0.5$, $\rho=1$, and $\rho=1.5$. (d) LINEX loss function with $a=0.5$, $a=1$, and $a=1.5$. (e) Proposed G-loss function with $a=0.5$, $a=1$, $a=1.5$, and $a=2$.} 
    \label{fig:Loss-Figues}
 \end{figure*}

\subsection{Loss functions}
In this subsection, we review some relevant loss functions, chosen to provide motivation for the work presented in this paper. Further, we provide the visual representation of baseline loss function in Fig. \ref{fig:Loss-Figues}. 

\begin{enumerate}
\item \textbf{Pinball loss function:} To improve the efficacy of TSVM against noise, \citet{tanveer2019general} proposed TSVM with pinball loss function (Pin-GTSVM). The mathematical formulation of the pinball loss function is expressed as:
\begin{align}
\mathcal{L}_{pin}(r)=
\begin{cases}
r, & r > 0, \\
-\tau r, & r \leq 0, 
\end{cases}
\end{align}
where $r=1-yf(x)$ and $\tau \in \left[0,1\right]$. For $\tau=0$, it reduces to the hinge loss function. It is asymmetric, unbounded, and non-smooth.

\item \textbf{Huber loss function:} To enhance the robustness of TSVM, \citet{borah2020functional} incorporated the Huber loss function in to TSVM. It is a combination of quadratic and linear loss. The mathematical representation of the Huber loss function
is articulated as follows:
\begin{align}
\mathcal{L}_{Huber}(r)=
\begin{cases}
\frac{1}{2} r^2, & |r| \leq \theta, \\
\theta |r| - \frac{1}{2} {\theta}^2, & \text{otherwise}, 
\end{cases}
\end{align}
where $\theta$ is a trade-off parameter between quadratic and linear loss. It is symmetric, unbounded, and smooth.

\item \textbf{Correntropy-induced loss function:} To enhance the robustness of TSVM against outliers, \citet{zheng2021ctsvm} introduced the correntropy-induced loss into TSVM and proposed a robust TSVM model. The mathematical expression
for the correntropy-induced loss is given as:
\begin{align}
\mathcal{L}_{ce}(r)= \lambda \left[ 1-\exp\left(\frac{-r^2}{\rho^2}\right)  \right], ~\forall~ r \in \mathbb{R},   
\end{align}
where $\rho$ is the normalizing constant and $\lambda > 0$ is the loss parameter. It is symmetric, bounded, and smooth.

\item \textbf{LINEX loss function:} To advance TSVM against noise or outliers, \citet{si2023symmetric} incorprated the LINEX loss into TSVM. The mathematical representation of the
LINEX loss is as follows:
\begin{align}  
\mathcal{L}_{LINEX}(r)= \exp\left(ar\right)-ar-1, ~\forall~ r \in \mathbb{R},
\end{align}
where $a \neq 0$ is the loss parameter that controls the penalty for classified and misclassified samples.  \citet{si2023symmetric} utilized the linear subpart of LINEX loss and introduced the symmetric LINEX loss into TSVM. It is unbounded and smooth. 
\end{enumerate}
In addition to these, recent advancements in designing robust and smooth functions include RoBoSS loss \cite{akhtar2023roboss}, Wave loss \cite{akhtar2024advancing, quadir2024enhancing}, HawkEye loss \cite{akhtar2024hawkeye}, and so forth.
\section{Proposed Work}
In this section, we present a novel advancement, the guardian loss function (G-loss), designed to shield the supervised algorithm against outliers or noise. Then, we amalgamate the proposed guardian loss function into TSVM and propose a novel robust and smooth classifier coined GL-TSVM.

\subsection{Guardian loss function}
In this subsection, we introduce a novel loss function, the guardian loss (G-loss), designed to guide and fortify traditional algorithms against outliers and noise. It is meticulously designed to manifest asymmetry, boundedness, and smooth characteristics. The mathematical formulation of the G-loss function is as follows:
\begin{align} \label{guardian-loss}   
\mathcal{L}_{G}(r)&= \frac{r \{\exp(ar)-1\}}{1+r \{\exp(ar)-1\}}, \nonumber\\
&= 1-\frac{1}{1+r \{\exp(ar)-1\}}, ~\forall~ r \in \mathbb{R},
\end{align}
where $r=1-yf(x)$ and $a >0$ is the parameter that governs the asymmetry of the G-loss function. Fig. \ref{fig:fifth } showcases the visual representation of the G-loss function for varying values of $a$. Its asymmetric, bounded, and smooth nature serves to guide the algorithm in the right direction during the training process; hence, it is named the guardian loss. The asymmetric feature of it allows for distinct penalties for distinct samples based on their positioning relative to the proximal hyperplane. Its bounded nature allows it to impose a strict limit on the maximum loss for data points with significant deviations, thereby mitigating the influence of noise or outliers. Furthermore, the smoothness property empowers the utilization of gradient-based iterative algorithms for model optimization.

\subsection{Linear GL-TSVM}
Given a binary training dataset, the objective of linear GL-TSVM is to seek positive and negative hyperplanes as follows:
\begin{align}\label{linear twin hyperplanes}
f_{+}={u}_{+}^{\top} {x}+b_{+}=0 ~~\text { and } \quad f_{-}={u}_{-}^{\top} {x}+b_{-}=0,
\end{align}
where ${u}_{+}$, ${u}_{-}$ $\in$ $\mathbb{R}^{n}$
and $b_{+}$, $b_{-}$ $\in$ $\mathbb{R}$
are the model parameters. To obtain the hyperplanes (\ref{linear twin hyperplanes}), we formed the primal problem of linear GL-TSVM as follows:\\
(Linear GL-TSVM-1)
\begin{align} \label{linear wave TSVM-1}
& \min _{{u}_{+}, b_{+}, \zeta^{-}} \frac{1}{2} \sum_{i=1}^{l_{+}}\left({u}_{+}^{\top} {x}_i+b_{+}\right)^2+\frac{1}{2} C_1\left(\left\|{u}_{+}\right\|_2^2+b_{+}^2\right)+C_2 \sum_{j=1}^{l_{-}} \zeta_j^{-}, \nonumber \\
& \text {s.t. } \zeta_j^{-}=   1- \frac{1}{1+ \left(1+{u}_{+}^{\top} {x}_j+b_{+}\right)  \left[\exp \{a\left(1+{u}_{+}^{\top} {x}_j+b_{+} \right)\} - 1\right] }, \quad j=1, \ldots, l_{-},
\end{align}
(Linear GL-TSVM-2)
\begin{align} \label{linear wave TSVM-2}
& \min _{{u}_{-}, b_{-}, \zeta^{+}} \frac{1}{2} \sum_{j=1}^{l_{-}}\left({u}_{-}^{\top} {x}_j+b_{-}\right)^2+\frac{1}{2} C_3\left(\left\|{u}_{-}\right\|_2^2+b_{-}^2\right)+C_4 \sum_{i=1}^{l_{+}} \zeta_i^{+}, \nonumber \\ 
& \text {s.t. } \zeta_i^{+}=   1- \frac{1}{1+ \left(1-{u}_{-}^{\top} {x}_i-b_{-}\right) \left[\exp \{a\left(1-{u}_{-}^{\top} {x}_i-b_{-} \right)\} -1\right]  }, \quad i=1, \ldots, l_{+},
\end{align}
where ${\zeta}_{+}=\left(\zeta_1^{+}, \ldots, \zeta_{l_{+}}^{+}\right)^{\top} \in \mathbb{R}^{l_{+}}, {\zeta}_{-}=\left(\zeta_1^{-}, \ldots, \zeta_{l_{-}}^{-}\right)^{\top} \in$ $\mathbb{R}^{l_{-}}$. To be concise, we solely discuss the optimization problem (\ref{linear wave TSVM-1}), with the understanding that optimization problem (\ref{linear wave TSVM-2}) follows a similar structure. The objective function outlined in equation (\ref{linear wave TSVM-1}) comprises three distinct components. The first component aims to minimize the distance between the positive hyperplane and the positive instances. The second component, a regularization term, is included to adhere to the structural risk minimization principle. Lastly, the third component accounts for the cumulative penalty of all negative samples by leveraging the proposed G-loss function. Due to the non-convex nature of optimization problems (\ref{linear wave TSVM-1}) and (\ref{linear wave TSVM-2}), and utilizing their inherent smoothness property, we devised an iterative algorithm to solve them. Initially, we convert (\ref{linear wave TSVM-1}) and (\ref{linear wave TSVM-2}) into vector-matrix form as follows:
\begin{align} \label{optimizationproblem1}
\min _{{u}_1} Q_1\left({u}_1\right)=\frac{1}{2}\left\|{M}^{\top} {u}_1\right\|_2^2+\frac{1}{2} C_1\left\|{u}_1\right\|_2^2+C_2 \mathcal{L}_1\left({u}_1\right),
\end{align}
% and
\begin{align} \label{optimizationproblem2}
\min _{{u}_2} Q_2\left({u}_2\right)=\frac{1}{2}\left\|{N}^{\top} {u}_2\right\|_2^2+\frac{1}{2} C_3\left\|{u}_2\right\|_2^2+C_4 \mathcal{L}_2\left({u}_2\right),
\end{align}
where $\mathcal{L}_1\left({u}_1\right)=\sum_{j=1}^{l_{-}}
 1- \frac{1}{1+ \left(1+{N}_j^{\top} {u}_1\right) \left[\exp \{a\left(1+{N}_j^{\top} {u}_1 \right)\} -1\right]  },~ j=1, \ldots, l_{-}$; $\mathcal{L}_2\left({u}_2\right)=\sum_{i=1}^{l_{+}}
  1- \frac{1}{1+ \left(1-{M}_i^{\top} {u}_2\right) \left[ \exp \{a\left(1-{M}_i^{\top} {u}_2 \right)\} -1 \right]  },~ i=1, \ldots, l_{+}$. ${M}=\left[{X}_{+}, {e}_1\right]^{\top} \in \mathbb{R}^{(n+1) \times l_{+}}$, ${N}=\left[{X}_{-}, {e}_2\right]^{\top} \in \mathbb{R}^{(n+1) \times l_{-}}$ ; ${u}_1=\left[{u}_{+}^{\top}, b_{+}\right]^{\top} \in$ $\mathbb{R}^{n+1}$, ${u}_2=\left[{u}_{-}^{\top}, b_{-}\right]^{\top} \in \mathbb{R}^{n+1}$. ${M}_i^{\top}$ is the $i^{th}$ row of ${M}$ and ${N}_j^{\top}$ is the $j^{th}$ row of ${N}$.  Further, for simplification, we use $A_j$ and $B_i$ to represent $\left(1+{N}_j^{\top} {u}_1 \right)$ and $\left(1-{M}_i^{\top} {u}_2 \right)$, respectively.

In accordance with the optimality condition, we obtain the following:
\begin{align}
& \nabla Q_1\left({u}_1\right)=\left({M} {M}^{\top}+C_1 {I}\right) {u}_1+ \hat{{N}} {s}_1=0, \label{problem1}\\
& \nabla Q_2\left({u}_2\right)=\left({N} {N}^{\top}+C_3 {I}\right) {u}_2 - \hat{{M}} {s}_2=0, \label{problem2}
\end{align}
where $\hat{{M}}=\left[C_4  {M}_1, \ldots, C_4 {M}_{l_{+}}\right] \in \mathbb{R}^{(n+1) \times l_{+}},~ \hat{{N}}=\left[C_2 {N}_1, \ldots, C_4 {N}_{l_{-}}\right] \in \mathbb{R}^{(n+1) \times l_{-}}$. ${s}_1=\left[s_{11}, \ldots, s_{1 l_{-}}\right]^{\top} \in \mathbb{R}^{l_{-}}$,~ ${s}_2=\left[s_{21}, \ldots, s_{2 l_{+}}\right]^{\top} \in \mathbb{R}^{l_{+}};~ s_{1 j}= \frac{\exp\left(aA_j\right) \left(aA_j+1\right) -1 }{ \left[1+ A_j \{\exp\left(aA_j\right) -1 \}  \right]  ^2},~ j=1, \ldots, l_{-};~ s_{2 i}= \frac{\exp\left(aB_i\right) \left(aB_i+1\right) -1 }{ \left[1+ B_i \{\exp\left(aB_i\right) -1 \}  \right]  ^2},~i=1, \ldots, l_{+}$.

Now, we use equations (\ref{problem1}) and (\ref{problem2}) to formulate iterative expressions for problems (\ref{optimizationproblem1}) and (\ref{optimizationproblem2}) in the following manner:
\begin{align} 
&{u}_1^{t+1}=-\left({M} {M}^{\top}+C_1 {I}\right)^{-1} \hat{{N}} {s}_1^t, \label{iterativeequation1}\\
&{u}_2^{t+1}=\left({N} {N}^{\top}+C_3 {I}\right)^{-1} \hat{{M}} {s}_2^t. \label{iterativeequation2}
\end{align}
% % and
% \begin{align} \label{iterativeequation2}
% {u}_2^{t+1}=\left({N} {N}^{\top}+C_3 {I}\right)^{-1} \hat{{M}} {s}_2^t .
% \end{align}
Here, $t$ represents the index of iteration. The iterative procedure involves iterating through equations (\ref{iterativeequation1}) and (\ref{iterativeequation2}) until convergence is achieved. After obtaining the solutions,  we can proceed to find the pair of hyperplanes (\ref{linear twin hyperplanes}). 

To ascertain the class of a unseen sample ${\widetilde{x}} \in \mathbb{R}^n$, we use the following decision rule:
\begin{align}
\text{Class of}~ {\widetilde{x}}=\arg \min _{i=+,-} \frac{\left|{u}_{i}^{\top} {\widetilde{x}}+b_{i}\right|}{\left\|{u}_{i}\right\|}.
\end{align}

\subsection{Non-linear GL-TSVM}
For the non-linear case, we utilized the kernel trick to map the input data points to a higher dimensional space. The objective of non-linear GL-TSVM is to identify a pair of hypersurfaces of the following form:
\begin{align} \label{hypersurface for nonlinear twin wave}
g_{+}=\mathcal{\kappa}\left({x}, {X}^{\top}\right) {v}_{+}+b_{+}=0 \quad \text { and } \quad g_{-}=\kappa\left({x}, {X}^{\top}\right) {v}_{-}+b_{-}=0,
\end{align}
where ${X}^{\top}=\left[{X}_{+} ; {X}_{-}\right]$ and $\kappa$ is the kernel function.

To determine the hypersurfaces (\ref{hypersurface for nonlinear twin wave}), we formulate the following optimization problems:\\
(Non-linear GL-TSVM-1)
\begin{align} \label{Non-linear Wave-TSVM-1}
& \min _{{v}_{+}, b_{+}, \zeta^{-}} \sum_{i=1}^{l_{+}} \frac{1}{2}\left(\kappa\left({x}_i, {X}^{\top}\right) {v}_{+}+b_{+}\right)^2+\frac{1}{2} C_1\left(\left\|{v}_{+}\right\|_2^2+b_{+}^2\right)+C_2 \sum_{j=1}^{l_{-}} \zeta_j^{-}, \nonumber \\
& \text {s.t. } \quad \zeta_j^{-}=  1- \frac{1}{1+ \left(1+\kappa\left({x}_j, {X}^{\top}\right) {v}_{+}+b_{+}\right) \left[   \exp \{a\left(1+\kappa\left({x}_j, {X}^{\top}\right) {v}_{+}+b_{+} \right)\} -1   \right]    }, \nonumber\\
& \quad j=1, \ldots, l_{-},
\end{align}
(Non-linear GL-TSVM-2)
\begin{align} \label{Non-linear Wave-TSVM-2}
&\min _{{v}_{-}, b_{-}, \zeta^{+}} \sum_{j=1}^{l_{-}} \frac{1}{2}\left(\kappa\left({x}_j, {X}^{\top}\right) {v}_{-}+b_{-}\right)^2+\frac{1}{2} C_3\left(\left\|{v}_{-}\right\|_2^2+b_{-}^2\right)+C_4 \sum_{i=1}^{l_{+}} {\zeta}_i^{+}, \nonumber\\
&\text {s.t. } \quad \zeta_i^{+}=  1- \frac{1}{1+ \left(1-\kappa\left({x}_i, {X}^{\top}\right) {v}_{-}-b_{-}\right) \left[\exp \{a\left(1-\kappa\left({x}_i, {X}^{\top}\right) {v}_{-}-b_{-} \right)\} -1 \right]}, \nonumber\\
&\quad i=1, \ldots, l_{+}.
\end{align}
% The method for solving problems (\ref{Non-linear Wave-TSVM-1}) and (\ref{Non-linear Wave-TSVM-2}) is akin to the linear case. 
The iterative method to solve (\ref{Non-linear Wave-TSVM-1}) and (\ref{Non-linear Wave-TSVM-2}) can be derived as follows:
\begin{align} \label{iterative equation1 nonlinear}
{v}_1^{{t}+1}= &- \left({G G}^{\top}+C_1 {I}\right)^{-1} \nonumber\\
&\left(\sum_{j=1}^{l_{-}} C_2 {H}_j \frac{  \exp\{a\left(1+{H}_j^{\top}{v}_1^t\right)\}   \{a\left(1+{H}_j^{\top}{v}_1^t\right)+1\} -1 }  
{\left[1+ \left(1+{H}_j^{\top}{v}_1^t\right) \{\exp \{a\left(1+{H}_j^{\top}{v}_1^t\right)\} -1\}
  \right]^2} 
  \right),
\end{align}
% and
\begin{align} \label{iterative equation2 nonlinear}
{v}_2^{t+1}= & \left({H} {H}^{\top}+C_3 {I}\right)^{-1} \nonumber\\
& \left(\sum_{i=1}^{l_{+}} C_4 {G}_i
\frac{  \exp\{a\left(1-{G}_i^{\top}{v}_2^t\right)\}   \{a\left(1-{G}_i^{\top}{v}_2^t\right)+1\} -1 }  
{\left[1+ \left(1-{G}_i^{\top}{v}_2^t\right) \{\exp \{a\left(1-{G}_i^{\top}{v}_2^t\right)\} -1\}
  \right]^2} 
  \right).
\end{align}

Here ${G}=\left[\kappa\left({X}_{+}, {X}^{\top}\right), {e}_1\right]^{\top} \in \mathbb{R}^{(l+1) \times l_{+}}$ , ${H}=\left[\kappa\left({X}_{-}, {X}^{\top}\right), {e}_2\right]^{\top} \in \mathbb{R}^{(l+1) \times l_{-}}$; ${G}_i$ is the $i^{th}$ column of the matrix ${G}$, ${H}_j$ is the $j^{th}$ column of the matrix ${H}$. ${v}_1=\left[{v}_{+}^{\top}, b_{+}\right]^{\top}$, ${v}_2=\left[{v}_{-}^{\top}, b_{-}\right]^{\top}$. Further, for simplification, we use $E_j$ and $F_i$ to represent $\left(1+{H}_j^{\top} {v}_1 \right)$ and $\left(1-{G}_i^{\top} {v}_2 \right)$, respectively.

It's important to highlight that equations (\ref{iterative equation1 nonlinear}) and (\ref{iterative equation2 nonlinear}) involve complex matrix inversions. Therefore, to alleviate computational complexity, we utilized the Sherman Morrison Woodbury theorem \cite{kumar2009least}. Subsequently, in equations (\ref{iterative equation1 nonlinear}) and (\ref{iterative equation2 nonlinear}), the inverse matrices are substituted with the following matrices:
% \begin{align} \label{new inverse matrix 1}
% {P_1}=\frac{1}{C_1}\left(I-{G}\left(C_1 {I}+{G}^{\top} {G}\right)^{-1} {G}^{\top}\right),
% \end{align}
% % and
% \begin{align}\label{new inverse matrix 2}
% {P_2}=\frac{1}{C_3}\left({I}-{H}\left(C_3 {I}+{H}^{\top} {N}\right)^{-1} {H}^{\top}\right) .
% \end{align}
\begin{align}
&{P_1}=\frac{1}{C_1}\left(I-{G}\left(C_1 {I}+{G}^{\top} {G}\right)^{-1} {G}^{\top}\right), \label{new inverse matrix 1}\\
& {P_2}=\frac{1}{C_3}\left({I}-{H}\left(C_3 {I}+{H}^{\top} {N}\right)^{-1} {H}^{\top}\right).\label{new inverse matrix 2}
\end{align}
Using the equations (\ref{new inverse matrix 1}) and (\ref{new inverse matrix 2}), the iterative approach can be derived in the following manner:
\begin{align}
&{v}_1^{{t}+1}=- P_1 \hat{{H}} {s}_1^t,\label{updated iterative equation1 nonlinear}\\
& {v}_2^{t+1}= P_2 \hat{{G}} {s}_2^t. \label{updated iterative equation2 nonlinear}
\end{align}
% % and
% \begin{align} \label{updated iterative equation2 nonlinear}
% {v}_2^{t+1}= & P_2 \hat{{G}} {s}_2^t.
% \end{align}
Here, $\hat{{H}}=\left[C_2 {H}_1, \ldots, C_2  {H}_{l_{-}}\right] \in \mathbb{R}^{(l+1) \times l_{-}}$, and $\hat{{G}}=\left[C_4 {G}_1, \ldots, C_4 {G}_{l_{+}}\right]$ $\in \mathbb{R}^{(l+1) \times l_{+}}$. Additionally, ${s}_1^t \in \mathbb{R}^{l_{-}}$, $s_{1j}^t= \frac{\exp\left(aE_j^{t}\right)\left(aE_j^{t}+1\right) -1 }         
{  \left[ 1 + E_j^{t} \{\exp\left(aE_j^{t}\right) -1 \}    \right]^2   }$, $j=1, \ldots, l_{-}$ ; ${s}_2^t \in \mathbb{R}^{l_{+}}$, $s_{2i}^t=\frac{\exp\left(aF_i^{t}\right)\left(aF_i^{t}+1\right) -1 }         
{  \left[ 1 + F_i^{t} \{\exp\left(aF_i^{t}\right) -1 \}    \right]^2   }$, $i=1, \ldots, l_{+}$. The iteration procedure is established by repeatedly applying equations (\ref{updated iterative equation1 nonlinear}) and (\ref{updated iterative equation2 nonlinear}) until convergence is reached. Consequently, upon obtaining the solutions ${v}_{+}, b_{+}$ and ${v}_{-}, b_{-}$, we can then determine the positive and negative hypersurfaces generated by the kernel.

To predict the class of a new sample ${\widetilde{x}} \in \mathbb{R}^n$, we use the following decision function:
% \begin{align}
% \text {Class of }~ {\widetilde{x}}= \begin{cases}+1, & \text { if } \frac{\left|\kappa\left({\widetilde{x}}, {X}^{\top}\right) {v}_{+}+b_{+}\right|}{\sqrt{{v}_{+}^{\top} \kappa\left({X}, {X}^{\top}\right) {v}_{+}}} \leq \frac{\left|\kappa\left({\widetilde{x}}, {X}^{\top}\right) {v}_{-}+b_{-}\right|}{\sqrt{{v}_{-}^{\top} \kappa\left({X}, {X}^{\top}\right) {v}_{-}}}, \\ -1, & \text { otherwise. }\end{cases}
% \end{align}
\begin{align}
\text{Class of}~ {\widetilde{x}}=\arg \min _{i=+,-} \frac{\left|\kappa\left({\widetilde{x}}, {X}^{\top}\right) {v}_{i}+b_{i}\right|}{\sqrt{{v}_{i}^{\top} \kappa\left({X}, {X}^{\top}\right) {v}_{i}}}.
\end{align}

The iterative algorithm structure for non-linear GL-TSVM subproblem (\ref{Non-linear Wave-TSVM-1}) is clearly described in Algorithm \ref{algorithm1}. The structure for subproblem (\ref{Non-linear Wave-TSVM-2}) is similar to it.
\begin{algorithm}
  \caption{Non-linear GL-TSVM}
  \label{algorithm1}
   \begin{algorithmic}
  \STATE \textbf{Input:}
     \STATE Training dataset:  $\left\{x_i,y_i\right\}_{i=1}^l$, $y_i \in\{-1,1\}$;
     \STATE The parameters: Convergence precision ($\eta$), maximum iteration number ($T$), parameter $C_1$ and $C_2$, G-loss parameter $a$, iteration number $t=0$;    
\STATE Initialize: $v_1^{0}$;
\STATE \textbf{Output:} $v_{+}$, $b_{+}$;  
\STATE $1:$ ${G}=\left[\kappa\left({X}_{+}, {X}^{\top}\right), {e}_1\right]^{\top}$ , ${H}=\left[\kappa\left({X}_{-}, {X}^{\top}\right), {e}_2\right]^{\top}$.
\STATE $2:$  \textbf{while} $t \leq T$ 
\STATE $3:$  ~~~~~\textbf{for} $j \leftarrow 1$ to $l_{-}$ 
\STATE $4:$ ~~~~~~~~~$s_{1j}^t  \leftarrow \frac{\exp\left(aE_j^{t}\right)\left(aE_j^{t}+1\right) -1 }         
{  \left[ 1 + E_j^{t} \{\exp\left(aE_j^{t}\right) -1 \}    \right]^2}$
\STATE $5:$  ~~~~~\textbf{end for}
\STATE $6:$  ~~~~~${v}_1^{{t}+1} \leftarrow - P_1 \hat{{H}} {s}_1^t$
\STATE $7:$  ~~~~~\textbf{if }$\left\|{v}_1^{t+1}-{v}_1^t\right\|<\eta$ 
\STATE $8:$ ~~~~~~~\text{break}
\STATE $9:$  ~~~~~\textbf{else}
\STATE $10:$  ~~~~~~~$t \leftarrow t+1$
\STATE $11:$  ~~~~\textbf{end if}
\STATE $12:$  \textbf{end while}
\STATE $13:$  $\left({v}_{+}^{\top}, b_{+}\right)^{\top}={v}_1^{t+1}$.
 \end{algorithmic}
\end{algorithm}
\subsection{Computational complexity}
Let $l$ and $n$ denote the number of samples and features in training dataset, respectively, and  $l_{+}$ and $l_{-}$ denote the count of positive and negative samples, respectively. The computational complexity of GL-TSVM primarily arises from the computation of matrix inversion. In the linear case, the algorithm requires solving the inverse of a matrix of order $(n + 1) \times (n + 1)$, which results in a time complexity of $\mathcal{O}((n + 1)^3)$. Consequently, the computational complexity of the proposed linear GL-TSVM is $\mathcal{O}(2T(n + 1)^3)$, where $T$ represents the maximum number of iterations. For the non-linear case, the algorithm needs to compute the inverse of two matrices: one of size $l_{+} \times l_{+}$ and the other of size $l_{-} \times l_{-}$, with computational complexities of $\mathcal{O}(l_{+}^3)$ and $\mathcal{O}(l_{-}^3)$, respectively.  Hence, for non-linear GL-TSVM, the computational complexity is $\mathcal{O}(T (l_{+}^3 + l_{-}^3))$. It is evident that non-linear GL-TSVM is not well-suited for large-scale problems due to its cubic computational complexity growth with respect to the size of positive and negative sample matrices. However, in future research, one can utilize the concept of granular computing to reduce the size of the sample matrices, making the non-linear GL-TSVM approach more feasible and effective for large-scale applications \cite{10619989}.

\section{Experimental Evaluation} To validate the effectiveness of the proposed GL-TSVM model, we evaluate it on 25 UCI \cite{dua2017uci} and KEEL \cite{derrac2015keel} benchmark datasets across various domains. For comparison, we used 6 state-of-the-art models, namely SVM \cite{cortes1995support}, TSVM \cite{khemchandani2007twin}, Pin-GTSVM \cite{tanveer2019general}, SLTSVM \cite{si2023symmetric}, Wave-TSVM \cite{akhtar2024advancing}, and IF-RVFL \cite{malik2022alzheimer}. Further, to showcase the efficacy of the proposed GL-TSVM in the biomedical realm, we evaluated it on the breast cancer (BreaKHis) and schizophrenia datasets. The detailed experimental setup employed for evaluating the models is provided in Section S.II of the supplementary file.

\subsection{Evaluation on UCI and KEEL datasets} 
For the linear case, the average classification accuracy of the proposed GL-TSVM and the baseline models are presented in Table \ref{tab:Linear-UCI-KEEL-Table}. The detailed experimental results for each dataset are presented in Table S.I of the supplementary file. The average accuracies of the existing SVM, TSVM, Pin-GTSVM, SLTSVM, and Wave-TSVM are $83.67 \%$, $86.99 \%$, $87.33 \%$, $84.18 \%$, and $87.13 \%$, respectively, whereas, the average accuracy of the proposed GL-TSVM is  $87.52\%$, surpassing the compared models. In terms of average accuracy, the proposed GL-TSVM secured the top position, while the Pin-GTSVM achieved the second position with an accuracy difference of $0.19$. Further, the average accuracy difference of proposed GL-TSVM from SVM, TSVM, SLTSVM, and Wave-TSVM are $3.85$, $0.53$, $3.34$, and $0.39$, respectively. This observation strongly underscores the competitiveness of the proposed linear GL-TSVM over baseline models. For the non-linear case, the average experimental results of the proposed GL-TSVM and baseline models are presented in Table \ref{tab:Non-linear-table}. The detailed experimental results for each dataset can be found in Table S.II of the supplementary file. The average accuracies of SVM, TSVM, Pin-GTSVM, IF-RVFL, SLTSVM, Wave-TSVM, and the proposed GL-TSVM are $85.36\%$, $88.45\%$, $88.51\%$, $82.57\%$, $89.17\%$, $89.97\%$, and $90.52\%$, respectively. Evidently, the proposed GL-TSVM achieves the highest classification accuracy with an accuracy difference of $0.55$ from the second-best model, Wave-TSVM. This finding firmly establishes the dominance of the non-linear GL-TSVM in comparison with the baseline models. 

To further support the efficacy of the proposed GL-TSVM, we performed a thorough statistical analysis using the rank test, Friedman test, Nemenyi post hoc test, and win-tie-loss test. A detailed discussion of the statistical tests and their results is presented in Section S.III of the supplementary file.

\begin{table}[t]
\centering
\caption{Average accuracy and rank of linear GL-TSVM against baseline models on benchmark UCI and KEEL datasets.}
\label{tab:Linear-UCI-KEEL-Table}
\resizebox{\textwidth}{!}{%
\begin{tabular}{lcccccc}
\hline
  & SVM \cite{cortes1995support} & TSVM \cite{khemchandani2007twin} & Pin-GTSVM \cite{tanveer2019general} & SLTSVM \cite{si2023symmetric} & Wave-TSVM \cite{akhtar2024advancing} & GL-TSVM$^{\dagger}$ \\
\hline
\textbf{Avg. Acc.} & 83.67 & 86.99 & \underline{87.33} & 84.18 & 87.13 & \textbf{87.52} \\ \hline
\textbf{Avg. Rank} & 4 & 3.64 & 3.22 & 4.4 & \underline{2.88} & \textbf{2.86} \\ \hline
\multicolumn{7}{l}{$^{\dagger}$ represents the proposed model.}\\
% \multicolumn{7}{l}{Here, Avg. and Acc. are acronyms used for average and accuracy, respectively.}\\
\multicolumn{7}{l}{The boldface and underline indicate the best and second-best models, respectively.} 
\end{tabular}}
\end{table}

\begin{table}[t]
\centering
\caption{Average accuracy and rank of non-linear GL-TSVM against the baseline models on benchmark UCI and KEEL datasets.}
\label{tab:Non-linear-table}
\resizebox{\textwidth}{!}{%
\begin{tabular}{lccccccc}
\hline
  & SVM \cite{cortes1995support} & TSVM \cite{khemchandani2007twin} & Pin-GTSVM \cite{tanveer2019general} & IF-RVFL \cite{malik2022alzheimer} & SLTSVM \cite{si2023symmetric} & Wave-TSVM \cite{akhtar2024advancing} & GL-TSVM$^{\dagger}$ \\
\hline
\textbf{Avg. Acc.} & 85.36 & 88.45 & 88.51 & 82.57 & 89.17 & \underline{89.97} & \textbf{90.52} \\ \hline
\textbf{Avg. Rank} & 5.26 & 4.08 & 3.9 & 5.92 & 3.68 & \underline{2.88} & \textbf{2.28} \\ \hline
\multicolumn{8}{l}{$^{\dagger}$ represents the proposed model.}\\
% \multicolumn{8}{l}{Here, Avg. and Acc. are acronyms used for average and accuracy, respectively.}\\
\multicolumn{8}{l}{The boldface and underline indicate the best and second-best models, respectively.}
\end{tabular}}
\end{table}

\subsection{Evaluation on breast cancer dataset}
To assess the effectiveness of the proposed GL-TSVM in practical scenarios, we evaluated it on breast cancer dataset named BreaKHis, accessible at \cite{spanhol2015dataset}. We utilized 1240 histopathology scans, each magnified 400 times. These scans predominantly fall into two categories: benign and malignant. Within the benign class, there are four subcategories: adenosis (ad), phyllodes tumor (pt), tubular adenoma (ta), and fibroadenoma (fd) with 106, 115, 130, and 237 scans, respectively. Similarly, the malignant class includes four subclasses: ductal carcinoma (dc), papillary carcinoma (pc), mucinous carcinoma (mc),and lobular carcinoma (lc) containing 208, 138, 169, and 137 scans, respectively. For feature extraction, we followed the same methodology as outlined in \cite{gautam2020minimum}.
The average results of the proposed GL-TSVM compared to the baseline models on the BreaKHis dataset are presented in Table \ref{tab:BreaKHis-Results-table}. The detailed performance comparison on each dataset is outlined in Table S.III of the supplement file. The average accuracy of the proposed GL-TSVM stands at $75.67\%$, surpassing the baseline models SVM, TSVM, Pin-GTSVM, IF-RVFL, SLTSVM, and Wave-TSVM which achieved accuracies of $65.15\%$, $74.14\%$, $74.21\%$, $64.7\%$, $72.48\%$, and $74.2\%$ respectively. Further, the average rank of the baseline models SVM, TSVM, Pin-GTSVM, IF-RVFL, SLTSVM, and Wave-TSVM is $6.22$, $2.91$, $3.19$, $6.19$, $4.38$, and $3.09$ while the average rank of GL-TSVM is $2.03$, representing the most favorable position in comparison to the baseline models.
 % Specifically, the average rank of the proposed GT-TSVM is $1.46$ lower than the second-best model, Pin-GTSVM, demonstrating a clear superiority in performance.} This indicates that the proposed GL-TSVM is more consistently accurate across different datasets than baseline TSVM based models.
 Further, the average rank differences between the proposed GL-TSVM and the other baseline models (SVM, TSVM, Pin-GTSVM, IF-RVFL, SLTSVM, and Wave-TSVM) are $4.19$, $0.88$, $1.46$, $4.16$, $2.35$ and $1.06$, respectively, showing substantial performance advantages for GL-TSVM over these models. Overall, the results showcase that the proposed GL-TSVM is significantly superior in the domain of breast cancer diagnosis when compared to the baseline models.

\subsection{Evaluation on schizophrenia dataset}
To further demonstrate the competitiveness of the proposed GL-TSVM model, we evaluated it for diagnosing schizophrenia patients. The data used in this study was obtained from the center for biomedical research excellence (COBRE) (\url{http://fcon_1000.projects.nitrc.org/indi/retro/cobre.html}). The dataset includes 72 schizophrenia subjects (ages 18-65, mean age 38.1 $\pm$ 13.9 years) and 74 healthy control subjects (ages 18-65, mean age 35.8 $\pm$ 11.5 years). The feature extraction process followed the methodology outlined in \cite{tanveer2022intuitionistic}. Table \ref{tab:Schezophrenia-Results-table} presents a comparative analysis of the performance of the proposed GL-TSVM model against baseline models. SVM has the lowest accuracy at 67.57\%. TSVM, Pin-GTSVM, and Wave-TSVM show improved accuracies of 75\%, 77.38\%, and 75\%, respectively, while IF-RVFL achieves 74.33\%. Notably, the SLTSVM model excels with an accuracy of 78.39\%, matched by the proposed GL-TSVM model. This indicates that the proposed GL-TSVM is as effective as the best-performing models, validating its efficacy as a reliable and competitive tool for schizophrenia diagnosis. 

\begin{table}[t]
\centering
\caption{Average performance of the proposed GL-TSVM against the baseline models on the BreaKHis
dataset.}
\label{tab:BreaKHis-Results-table}
\resizebox{\textwidth}{!}{%
\begin{tabular}{lccccccc}
\hline
 & SVM \cite{cortes1995support} & TSVM \cite{khemchandani2007twin} & Pin-GTSVM \cite{tanveer2019general} & IF-RVFL \cite{malik2022alzheimer} & SLTSVM \cite{si2023symmetric} & Wave-TSVM \cite{akhtar2024advancing} & GL-TSVM$^{\dagger}$ \\
\hline
\textbf{Avg. Acc.} & 65.15 & 74.14 & \underline{74.21} & 64.7 & 72.48 & 74.2 & \textbf{75.67} \\ \hline
\textbf{Avg. Rank} & 6.22 & \underline{2.91} & 3.19 & 6.19 & 4.38 & 3.09 & \textbf{2.03} \\ \hline
\multicolumn{8}{l}{$^{\dagger}$ represents the proposed model.}\\
% \multicolumn{8}{l}{Here, Avg. and Acc. are acronyms used for average and accuracy, respectively.}\\
\multicolumn{8}{l}{The boldface and underline indicate the best and second-best models, respectively.}
\end{tabular}}
\end{table}

%%% SF Table
\begin{table}[t]
\centering
\caption{Average performance of the proposed GL-TSVM against the baseline models on the schizophrenia
dataset.}
\label{tab:Schezophrenia-Results-table}
\resizebox{\textwidth}{!}{%
\begin{tabular}{lccccccc}
\hline
 & SVM \cite{cortes1995support} & TSVM \cite{khemchandani2007twin} & Pin-GTSVM \cite{tanveer2019general} & IF-RVFL \cite{malik2022alzheimer} & SLTSVM \cite{si2023symmetric} & Wave-TSVM \cite{akhtar2024advancing} & GL-TSVM$^{\dagger}$ \\
\hline
    \textbf{Acc.} & 67.57 & 75 & \underline{77.38} & 74.33 & \textbf{78.39} & 75 & \textbf{78.39} \\ \hline
    \multicolumn{8}{l}{$^{\dagger}$ represents the proposed model.}\\
% \multicolumn{8}{l}{Here, Avg. and Acc. are acronyms used for average and accuracy, respectively.}\\
\multicolumn{8}{l}{The boldface and underline indicate the best and second-best models, respectively.}
\end{tabular}
}
\end{table}
\subsection{Effectiveness of the G-Loss function}
The experimental findings affirm the efficacy of the G-loss function in bolstering the robustness and performance of GL-TSVM. The asymmetric design of the G-loss function permits differential handling of samples relative to their distance from the decision boundary, effectively reducing the influence of noise. Its bounded characteristic caps the maximum loss, thereby preventing extreme values from unduly affecting the model. Additionally, the smoothness of the G-loss function supports the implementation of gradient-based optimization techniques, facilitating both efficient and stable convergence. Collectively, these features significantly enhance the performance of the proposed GL-TSVM, as demonstrated by the experimental results across a diverse array of datasets and domains.
\section{Conclusions}
In conclusion, the introduction of the G-loss function and the development of GL-TSVM model have significantly enhanced the robustness and performance of traditional TSVM algorithms. By leveraging the asymmetry, boundedness, and smoothness properties of the G-loss function, the GL-TSVM model offers a novel approach to handling outliers or noise in the data. Furthermore, the inclusion of a regularization term to adhere to the structutal risk minimization principle led to the creation of a more powerful classifier in the form of GL-TSVM. The iterative algorithm utilized for optimizing the GL-TSVM model ensures efficient convergence and stability. The experimental evaluations on a diverse set of benchmark datasets have consistently demonstrated the superior performance of GL-TSVM compared to baseline models. The application of GL-TSVM to breast cancer (BreaKHis) and schizophrenia datasets further validates its effectiveness in the biomedical domain. 
\par
However, it is noteworthy that, due to the computation of matrix inversion, GL-TSVM is not well-suited for large-scale problems. In the future, one can reformulate the GL-TSVM to circumvent the need to compute matrix inversion. Further, in the future, researchers can explore the fusion of the G-loss function with cutting-edge methodologies like support matrix machines \cite{kumari2023support} to tackle complex real-world problems.

\section*{Acknowledgment}
This project receives support from the Science and Engineering Research Board through the Mathematical Research Impact-Centric Support (MATRICS) scheme, with Grant No. MTR/2021/000787. Additionally, Mushir Akhtar's research fellowship is provided by the Council of Scientific and Industrial Research (CSIR), New Delhi, under Grant No. 09/1022(13849)/2022-EMR-I.

\bibliography{refs.bib}
\bibliographystyle{abbrvnat}
\end{document}